\documentclass[11pt]{article}

\usepackage[final]{acl}

\usepackage{times}
\usepackage{latexsym}

\usepackage[T1]{fontenc}

\usepackage[utf8]{inputenc}

\usepackage{microtype}

\usepackage{inconsolata}

\usepackage{graphicx}

\usepackage{xcolor}
\usepackage{algorithm}
\usepackage{algpseudocode}
\usepackage{amsmath}
\usepackage{booktabs}
\usepackage{amssymb}

\title{DynaGen: Unifying Temporal Knowledge Graph Reasoning with Dynamic Subgraphs and Generative Regularization}

\author{
\textbf{Jiawei Shen}\textsuperscript{1}, 
\textbf{Jia Zhu}\textsuperscript{1}, 
\textbf{Hanghui Guo}\textsuperscript{1}, 
\textbf{Weijie Shi}\textsuperscript{2},\\
\textbf{Guoqing Ma}\textsuperscript{1}, 
\textbf{Yidan Liang}\textsuperscript{1}, 
\textbf{Jingjiang Liu}\textsuperscript{1}, 
\textbf{Hao Chen}\textsuperscript{3},
\textbf{Shimin Di}\textsuperscript{4}\\[0.5em] 
\textsuperscript{1}Zhejiang Normal University, Zhejiang, China\\
\textsuperscript{2}Hong Kong University of Science and Technology, Hong Kong, China\\
\textsuperscript{3}Tencent, China \quad\quad
\textsuperscript{4}Southeast University, Jiangsu, China
}

\begin{document}
\maketitle
\begin{abstract}
Temporal Knowledge Graph Reasoning (TKGR) aims to complete missing factual elements along the timeline. Depending on the temporal position of the query, the task is categorized into interpolation and extrapolation. Existing interpolation methods typically embed temporal information into individual facts to complete missing historical knowledge, while extrapolation techniques often leverage sequence models over graph snapshots to identify recurring patterns for future event prediction.
These methods face two critical challenges: limited contextual modeling in interpolation and cognitive generalization bias in extrapolation. To address these, we propose a unified method for TKGR, dubbed DynaGen. For interpolation, DynaGen dynamically constructs entity-centric subgraphs and processes them with a synergistic dual-branch GNN encoder to capture evolving structural context. For extrapolation, it applies a conditional diffusion process, which forces the model to learn underlying evolutionary principles rather than just superficial patterns, enhancing its ability to predict unseen future events. Extensive experiments on six benchmark datasets show DynaGen achieves state-of-the-art performance. On average, compared to the second-best models, DynaGen improves the Mean Reciprocal Rank (MRR) score by 2.61 points for interpolation and 1.45 points for extrapolation. 

\end{abstract}

\section{Introduction}
Temporal Knowledge Graphs (TKGs) systematically characterize the dynamic evolution of relations between entities over time \cite{lin2023tflex,chen2024decrl,wang2024llmda}. They have become pivotal in modeling dynamic interactions across recommendation systems \cite{wang2019kgat}, financial risk prediction \cite{sang2022dialoguegat}, and information retrieval \cite{liu2018entity}.
Temporal Knowledge Graph Reasoning (TKGR) builds upon TKGs as its foundational data structure, utilizing the temporal dynamics and relational patterns inherent in TKGs to complete missing events or forecast future events for specific timestamps \cite{wang2024llmda}. Specifically, TKGR tasks can be categorized into \textbf{interpolation reasoning} and \textbf{extrapolation reasoning} \cite{chen2024unified}. The former focuses on completing missing knowledge within historical time ranges, while the latter is dedicated to predicting unknown events in the future \cite{fang2024eceformer}. Although recent methods have achieved great success, they still struggle to comprehensively address the complexity of real-world TKGs \cite{xin2024kartgps,liu2023retia}, which we detail in the following two aspects.

\begin{figure}
  \centering
  \includegraphics[width=0.45\textwidth]{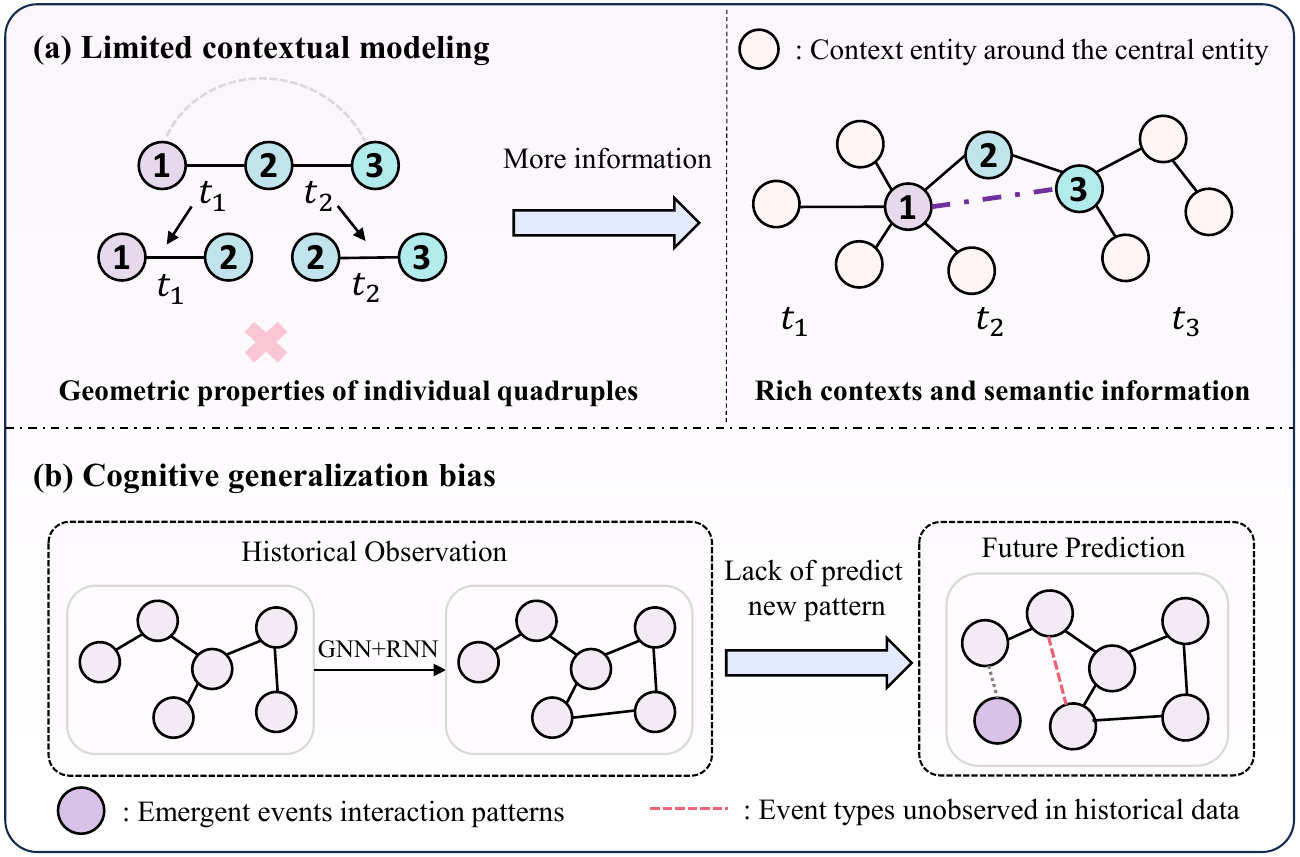}
  \caption{Traditional approaches suffer from: (a) Limited contextual modeling in interpolation reasoning; (b) Cognitive generalization bias in extrapolation reasoning.}
  \label{challenge}
\end{figure}

\textbf{Challenge 1: Limited contextual modeling in interpolation reasoning.} Existing interpolation methods lie in a modeling paradigm that treats temporal facts as relatively isolated units. These approaches often focus on learning the interaction patterns within individual quadruples $(s, p, o, t)$, achieving success by capturing local, pairwise semantics \cite{zhang2023hyie,ying2024tcompounde}. However, as illustrated in Figure~\ref{challenge}(a), this overlooks the crucial role of the \textbf{dynamic and evolving structure} centered around the query entity. By failing to capture this dynamic structural context, these models struggle to understand the complete logical chain behind complex events.

\textbf{Challenge 2: Cognitive generalization bias in extrapolation reasoning.} In open dynamic systems, future predictions must contend with emergent events, which are novel interaction patterns unobserved in historical data. However, prevailing extrapolation approaches adopt a two-stage pipeline, consisting of structural encoding for each time slice followed by RNN components for temporal dependency modeling \cite{chen2024decrl,tang2024gtrl,xu2023cenet}. As depicted in Fig.~\ref{challenge}(b), such pipelines inherently bias predictions toward the empirical training distribution and omit explicit modeling of the underlying generative principles. In essence, these models learn to imitate observed patterns rather than understanding the \textbf{underlying generative mechanism} that drives the evolution of the knowledge graph. When faced with future events spawned from novel dynamics, their predictive power collapses.

To address the above challenges, we present \textbf{DynaGen}, a unified method for TKGR with \textbf{Dyna}mic Subgraphs and \textbf{Gen}erative Reasoning. The core design of DynaGen features two synergistic modules, each engineered to tackle a specific challenge in TKGR:

\textbf{(1) Synergistic Structural Encoding for Interpolation.}
To overcome the limitations of contextual modeling, we first dynamically construct a temporally-aware subgraph for each query, which captures a rich structural context that goes far beyond analyzing isolated facts. This subgraph is then processed by \textbf{Synergistic Structure-Aware Encoder (SSAE)}, a dual-branch Graph Neural Network. By integrating a Relational Graph Convolutional Network for relation semantics and a Graph Attention Network for structural importance, this encoder captures multi-faceted context far beyond any single fact, which is critical for accurate interpolation.

\textbf{(2) Generative Regularization for Extrapolation.}
To mitigate generalization bias in extrapolation, we introduce a new training strategy using a diffusion process as a conditional regularizer. Uniquely, we apply this generative process to the output of SSAE. By training the model to reconstruct this structural representation from only the query relation and time, we compel it to learn the \textbf{underlying generative knowledge} of temporal evolution, not just superficial patterns. This robust representation is then passed to a Transformer-Mixer architecture for final reasoning, significantly enhancing the model's ability to generalize to unseen future events. Our main contributions are as follows:
\begin{itemize}
    \item We dynamically construct temporally-aware subgraphs for each query and propose a synergistic dual-branch encoder that models these subgraphs to overcome the limitations of contextual modeling.
    \item We introduce a new generative regularization strategy, applying a diffusion process to learn the underlying principles of temporal evolution. This approach, combined with a unified reasoning module, significantly enhances the model's extrapolation capabilities and mitigates generalization bias in extrapolation.
    \item DynaGen unifies interpolation and extrapolation in a single end‑to‑end framework and achieves consistent state‑of‑the‑art results on six TKG datasets.
\end{itemize}

\section{Related Work}

\subsection{Interpolation Reasoning}
Interpolation reasoning aims to infer missing facts by embedding temporal information into the evolutionary trajectories of entities and relations \cite{li2023teast, pan2024hge, chen2024unified}. These methods are broadly categorized into two types. Timestamp-independent approaches, such as TTransE \cite{leblay2018deriving} and BoxTE \cite{messner2022boxte}, treat time as a discrete attribute. While efficient, they often fail to capture complex temporal dynamics. In contrast, timestamp-specific methods employ dedicated time-aware functions for more effective modeling. Notable examples include RotateQVS \cite{chen2022rotateqvs}, which uses temporal rotations, HyIE \cite{zhang2023hyie}, which leverages hybrid geometric spaces, and LCGE \cite{Niu2023lcge}, which models event timeliness and causality. 
While these approaches have advanced the field by capturing the local, pairwise semantics within individual facts, they adhere to a modeling paradigm that overlooks a crucial element: the dynamic structural context. By treating events as relatively isolated, they struggle to understand the complete logical chain that emerges from the evolving structure around a query entity.
\begin{figure*}[t]
  \centering
  \includegraphics[width=\linewidth]{framework.pdf}
  \caption{The framework of DynaGen. DynaGen first constructs a dynamic centric subgraph. This subgraph is then processed by a dual-branch GNN encoder to generate an initial embedding. Next, encoded embedding is regularized via diffusion during training and refined by a Transformer-Mixer, leading to the final entity prediction. Finally, the score is calculated based on the final representation to obtain the rank.}
  \label{framework}
\end{figure*}
\subsection{Extrapolation Reasoning}
Extrapolation reasoning aims to forecast future events by identifying patterns in historical data \cite{chen2023mtkge, huang2024tempvalid}. Many innovative methods leverage Graph Neural Networks for this purpose \cite{chen2024decrl, chen2024logcl, chen2024unified}. Foundational approaches include R-GCN \cite{schlichtkrull2018rgcn}, which adapted convolutions for multi-relational graphs, and RENET \cite{jin2020renet}, which used an RNN to encode historical facts. More recent works have introduced specialized mechanisms. For example, L$^2$TKG \cite{zhang2023l2tkg} learns latent relations, CSI \cite{chen2024csi} extracts causal sub-histories to improve logical predictions, and LLM-DA \cite{wang2024llmda} dynamically updates rules generated by large language models. Other methods have also shown strong performance \cite{zheng2023dream, tang2024dhyper, cao2025dpcl_diff}. 
Despite these advancements, a prevailing paradigm in these methods involves a two-stage pipeline of structural encoding and temporal dependency modeling. This approach inherently restricts models to imitate patterns observed in the training data, omitting explicit modeling of the underlying generative mechanism. Consequently, they exhibit a cognitive generalization bias and are ill-equipped to predict emergent events.

\section{Methodology}
In this section, we propose the \textbf{DynaGen} for Temporal Knowledge Graph Reasoning (TKGR), as illustrated in Figure \ref{framework}.

\subsection{Preliminary}

\paragraph{Temporal Knowledge Graph (TKG).}
A collection of facts associated with timestamps. Formally, it is represented by a set of quadruples $\mathcal{Q} \subseteq \mathcal{E} \times \mathcal{R} \times \mathcal{E} \times \mathcal{T}$, where $\mathcal{E}$, $\mathcal{R}$, and $\mathcal{T}$ denote the sets of entities, relations, and timestamps, respectively. Each quadruple $(s, p, o, t) \in \mathcal{Q}$ represents a fact where subject $s$ and object $o$ are linked by relation $p$ at timestamp $t$.

\paragraph{Entity Prediction Task.}
Given a TKG, the goal is to complete a query with a missing entity. For example, the query $(s, p, ?, t)$ aims to predict the missing object, while $(?, p, o, t)$ aims to predict the missing subject.
This prediction is performed based on a sequence of observed graph snapshots $\{\mathcal{G}_1, \mathcal{G}_2, \dots, \mathcal{G}_T\}$, where $\mathcal{G}_t$ is the set of facts at timestamp $t$, and $T$ is the maximum observed time.
The task comprises two primary settings: \textbf{interpolation}, which predicts missing facts within the observed time range $[1, T]$, and \textbf{extrapolation}, which predicts future facts beyond time $T$.

\subsection{Adaptive Temporal Subgraph Construction}
\label{subsec:subgraph}

To capture a rich temporal context while managing computational complexity, we propose an adaptive entity-centric subgraph construction method. Given a query $(e_i, p_i, t_i)$, this module dynamically determines: (i) an appropriate time-window size $\Delta t$, and (ii) a temporal weight for each retained edge to construct a subgraph $\mathcal{G}_i$. This resulting subgraph is a weighted directed graph rooted at $e_i$, encoding both structural and temporal proximity.

\paragraph{Time-Window Prediction.}
A lightweight MLP takes the concatenation of the query entity embedding $\mathbf{e}_i\in\mathbb{R}^d$ and the query-time embedding $\mathbf{t}_i\in\mathbb{R}^d$, producing a scalar value $\delta\in[0,1]$:
\begin{equation}
\delta = \sigma\bigl(\,\mathbf{W}_2\;\text{ReLU}(\mathbf{W}_1
[\mathbf{e}_i;\mathbf{t}_i]+\mathbf{b}_1)+\mathbf{b}_2\,\bigr).
\end{equation}
The concrete time-window $\Delta t$ is then scaled linearly within a predefined range:
\begin{equation}
\Delta t = \delta \cdot(\tau_{\max}-\tau_{\min})+\tau_{\min},
\end{equation}
where $\tau_{\max}$ and $\tau_{\min}$ are hyperparameters that define the boundaries of the dynamic window size.

\paragraph{Multi-hop Temporal Neighbor Expansion.}
Starting from the query entity $e_i$, we perform a breadth-first search (BFS) within the dynamically computed time interval $[t_i - \Delta t, t_i]$. For each entity $v$ explored during the search, we collect its \emph{time-filtered} neighbors:
\begin{equation}
\label{eq:neighbor_collection}
\begin{split}
    \mathcal{N}_\Delta(v) = \left\{ (v', \tilde{r}, t) \mid (v, r, v', t) \in \mathcal{G} \right. \\ \left. \vee (v', r, v, t) \in \mathcal{G}, t \in [t_i - \Delta t, t_i] \right\},
\end{split}
\end{equation}
where inverse relations $\tilde{r} = r + |\mathcal{R}|$ are introduced. The BFS expansion terminates when the hop count reaches a maximum of $K_{\max}$.

\paragraph{Temporal Edge Weighting.}
To emphasize more recent facts, each retained edge $e$ discovered during the BFS is assigned an exponentially decaying weight:
\begin{equation}
w(e) = \exp\left(-\gamma \, |t_i - t|\right), \quad \gamma > 0,
\end{equation}
where $\gamma$ is a learnable scalar.

\paragraph{Final Subgraph Assembly.}
The final subgraph $\mathcal{G}_i = (\mathcal{V}_i, \mathcal{E}_i)$ is formed by aggregating all nodes and edges discovered during the multi-hop expansion. The node set $\mathcal{V}_i$ comprises the root entity $e_i$ and all unique entities visited by the BFS, while the edge set $\mathcal{E}_i$ consists of the corresponding time-filtered, weighted edges. This subgraph $\mathcal{G}_i$ is then passed to the downstream encoder (Section~\ref{subsec:gnn}). More details of the subgraph construction are provided in Appendix~\ref{sec:appendix_subgraph_algo}.

\subsection{Synergistic Structure-Aware Encoder}
\label{subsec:gnn}

Encoding the constructed subgraph $\mathcal{G}_i$ presents a multi-faceted challenge: the model must simultaneously interpret \textbf{relation semantics} (what type of connection), infer \textbf{structural importance} (which neighbors are more salient), and understand \textbf{temporal dynamics} (when did events happen). A single, monolithic GNN architecture often struggles to address all aspects effectively.
To tackle this, we propose a synergistic structure-aware encoder (\textbf{SSAE}) with a \textbf{dual-branch architecture} that explicitly disentangles and then integrates these different facets of information. 

\paragraph{Branch 1: Modeling Relation Semantics.}
To capture the explicit meaning of different relation types, which is crucial for logical reasoning, this branch employs a Relational Graph Convolutional Network (R-GCN) layer. It generates a relation-aware message $\mathbf{m}_{v}^{\text{(R-GCN)}}$ by aggregating information from neighbors through type-specific transformations:
\begin{equation}
\mathbf{m}_{v}^{\text{(R-GCN)}} = \sum_{r \in \mathcal{R}} \sum_{u \in \mathcal{N}_r(v)} \frac{1}{|\mathcal{N}_r(v)|} \mathbf{W}_r^{(l)} \mathbf{h}_u^{(l)}.
\end{equation}

\paragraph{Branch 2: Capturing Structural Importance.}
However, the relation is not the sole indicator of a neighbor's importance. To learn a more flexible, context-dependent weighting of neighbors, we run a parallel Graph Attention Network (GAT) branch. This branch computes a structural message $\mathbf{m}_{v}^{\text{(GAT)}}$ by assigning attention scores to neighbors based on their features, thus identifying the most salient nodes in the local structure:
\begin{equation}
\mathbf{m}_{v}^{\text{(GAT)}} = \bigoplus_{k=1}^{K} \sigma \left( \sum_{u \in \mathcal{N}(v)} \alpha_{uv,k}^{(l)} \mathbf{W}_{k}^{(l)} \mathbf{h}_u^{(l)} \right).
\end{equation}

\paragraph{Temporal Gating, Fusion, and Update.}
The information from both branches is then refined based on temporal recency. We introduce a \textbf{temporal gate} $g_{uv} \in [0, 1]$ that modulates the strength of each message based on its temporal distance from the query time $t_i$. The messages from the two branches are first gated, then concatenated and fused by a linear layer:
\begin{equation}
\mathbf{m}_{v}^{\text{(fused)}} = \mathbf{W}_{\text{fuse}}^{(l)} \left[ g_{uv} \cdot \mathbf{m}_{v}^{\text{(R-GCN)}}; g_{uv} \cdot \mathbf{m}_{v}^{\text{(GAT)}} \right].
\end{equation}
Finally, the fused message is integrated with the node's previous representation via a residual connection to produce the updated feature for the next layer:
\begin{equation}
\mathbf{h}_v^{(l+1)} = \sigma \left( \mathbf{W}_{\text{self}}^{(l)}\mathbf{h}_v^{(l)} + \mathbf{m}_{v}^{\text{(fused)}} \right).
\end{equation}

\paragraph{Final Query Representation.}
After $L$ such layers, we take the final representation of the root node, $\mathbf{z}_i = \mathbf{h}_{e_i}^{(L)}$. This output representation is now enriched with complementary semantic, structural, and temporal information, making it a robust input for the subsequent diffusion regularization and final prediction stages.

\subsection{Diffusion-based Output Regularization}
\label{subsec:diffusion}

To ensure DynaGen learns the underlying generative principles of TKG evolution, rather than simply memorizing observed patterns, we introduce a conditional diffusion process. 
This process is aim to teach the model a general rule for how local structures form around a given event.

The SSAE encoder produces $z_i$ that represents the observed neighborhood of a query. We then train a diffusion model to reconstruct $z_i$ conditioned only on the essential query information: the relation $p_i$ and time $t_i$. By tasking the model with generating a complete structural representation from such minimal input, we force it to learn a generalizable principle: what the context for this type of event should look like. This generative training objective ensures representations capture the fundamental dynamics of the TKG, significantly enhancing the model's ability to generalize to unseen events.

\paragraph{Forward Diffusion Process.}
During training, we corrupt the SSAE's output $\mathbf{z}_i$ by iteratively adding Gaussian noise over $K$ discrete timesteps. This forward process is defined as:
\begin{equation}
q(\mathbf{z}_k \mid \mathbf{z}_{k-1}) = \mathcal{N}(\mathbf{z}_k; \sqrt{1-\beta_k}\mathbf{z}_{k-1}, \beta_k\mathbf{I}),
\end{equation}
where $k \in \{1, \dots, K\}$ is the diffusion step, and $\{\beta_k\}_{k=1}^K$ is a predefined variance schedule. A closed-form solution allows us to sample the noised representation $\mathbf{z}_k$ at any step $k$ directly from the original $\mathbf{z}_i$:
\begin{equation}
    \mathbf{z}_k = \sqrt{\bar{\alpha}_k}\mathbf{z}_i + \sqrt{1 - \bar{\alpha}_k}\boldsymbol{\epsilon},
\end{equation}
where $\boldsymbol{\epsilon} \sim \mathcal{N}(\mathbf{0}, \mathbf{I})$ is the injected noise, $\alpha_k = 1 - \beta_k$, and $\bar{\alpha}_k = \prod_{i=1}^{k} \alpha_i$.

\paragraph{Conditional Denoising Network.}
We then train a denoising network, $\boldsymbol{\epsilon}_\theta$, to predict the noise $\boldsymbol{\epsilon}$ that was added to $\mathbf{z}_i$, based on the corrupted vector $\mathbf{z}_k$, the diffusion step $k$, and the core query information. This makes the denoising process conditional. The conditioning information is the concatenation of the query relation embedding $\mathbf{p}_i$ and the query time embedding $\mathbf{t}_i$. The denoising network is formulated as:
\begin{equation}
\begin{split}
\hat{\boldsymbol{\epsilon}} &= \boldsymbol{\epsilon}_\theta(\mathbf{z}_k, k, \mathbf{p}_i, \mathbf{t}_i) \\ &= \text{MLP}([\mathbf{z}_k; \text{PE}(k); \mathbf{p}_i; \mathbf{t}_i]),
\end{split}
\end{equation}
where $\text{PE}(\cdot)$ is the sinusoidal positional embedding for the diffusion step.

\paragraph{Training Objective and Inference.}
The denoising network is trained via a simple mean-squared error loss, which forms an auxiliary objective during the main training loop:
\begin{equation}
    \mathcal{L}_{\text{diff}} = \mathbb{E} \left\| \boldsymbol{\epsilon} - \boldsymbol{\epsilon}_\theta(\mathbf{z}_k, k, \mathbf{p}_i, \mathbf{t}_i) \right\|^2.
\end{equation}
This process acts as a powerful regularizer, compelling the SSAE to generate representations $\mathbf{z}_i$ that are robust and highly informative with respect to the query context.

Crucially, the entire diffusion and denoising module is \textbf{only active during training}. At inference time, it is completely discarded. We use the original, clean output $\mathbf{z}_i$ from the SSAE directly for the final prediction, incurring no additional computational cost.

\subsection{Unified Reasoning over the Local Neighborhood}
\label{subsec:transformer}
While the above module provides multi-hop structural information, 
making accurate temporal predictions often requires more fine-grained reasoning. Furthermore, models are prone to learning statistical "shortcuts" from training data, which may not generalize well when extrapolating to future, unseen events.

To address these issues, we propose a two-stage unified reasoning module built upon Transformer and MLP-Mixer architectures. This module is to first model the internal structure of each individual fact at the "atomic" level, and then efficiently aggregate information across the entire sequence of contextual facts at the "global" level. 

\paragraph{Input Sequence Construction.}
For a given query $(e_i, p_i, t_i)$ and its enhanced representation $\mathbf{z}_i$, we construct an input sequence. This sequence begins with a special query token $C_0$, where we mask the timestamp to introduce an auxiliary self-supervised task. This token is followed by the set of $k$ 1-hop neighboring facts $\mathcal{C} = \{(e_i, p_i, t_i)\}_{i=1}^k$ extracted during the subgraph construction phase.
\begin{equation}
    \text{Query Token } C_0 = [\texttt{[CLS]}, \mathbf{z}_i, \mathbf{p}_i, \texttt{[MASK]}],
\end{equation}
\begin{equation}
    \text{Input Sequence } \mathcal{S} = (C_0, C_1, \dots, C_k).
\end{equation}

Each element in the sequence is embedded, augmented with positional encodings, and fed into the two-stage encoder.

\paragraph{Stage 1: Atomic Fact Encoding.}
The first stage uses a standard Transformer encoder, $\mathcal{F}_{\text{atomic}}$, to capture fine-grained, local dependencies. Its self-attention mechanism excels at modeling the internal relationships within each fact representation as well as the interactions between the query token and its immediate neighbors. This produces a sequence of contextually-aware hidden states, $\mathbf{H} = \mathcal{F}_{\text{atomic}}(\mathcal{S})$.

\paragraph{Stage 2: Global Context Aggregation.}
While another Transformer could process $\mathbf{H}$, doing so for a potentially long sequence of facts is computationally expensive. To efficiently aggregate information across the entire sequence, we employ a parameter-efficient MLP-Mixer architecture, $\mathcal{F}_{\text{global}}$. The Mixer consists of $L/2$ layers, each alternating between: (i) token-mixing MLPs, applying MLPs across the sequence dimension, and (ii) channel-mixing MLPs, applying MLPs across the feature dimension. This design allows for effective global communication across all facts in the context.

The final representation $\mathbf{z}_{\text{final}}$ is obtained by mean-pooling the output of the MLP-Mixer. This representation, which has integrated multi-hop structural awareness, fine-grained atomic relationships, and global contextual information, is then used for the final entity prediction.

\subsection{Prediction and Training Objective}

\paragraph{Entity Prediction.}
The final representation $\mathbf{z}_{\text{final}}$ produced by the unified reasoning module encapsulates the rich structural, semantic, and temporal context of the query. We use this representation to score all potential candidate entities $e_j \in \mathcal{E}$ via a simple dot product similarity:
\begin{equation}
\text{score}(e_j) = \mathbf{z}_{\text{final}} \cdot \mathbf{e}_j^\top,
\end{equation}
where $\mathbf{e}_j$ is the embedding of the candidate entity $e_j$. During inference, this is efficiently computed for the entire entity vocabulary $\mathcal{E}$ using a single matrix multiplication, $\mathbf{s} = \mathbf{z}_{\text{final}} \mathbf{E}^\top$, where $\mathbf{E} \in \mathbb{R}^{|\mathcal{E}| \times d}$ is the full entity embedding matrix.

\paragraph{Training Objective.}
Our model is trained end-to-end by optimizing a composite loss function that combines the primary entity prediction task with two auxiliary regularization tasks. The main prediction loss, $\mathcal{L}_{\text{pred}}$, is the standard cross-entropy loss over the entity scores, enhanced with label smoothing (LS) for regularization:
\begin{equation}
\mathcal{L}_{\text{pred}} = - \sum_{j=1}^{|\mathcal{E}|} y_j^{\text{LS}} \log \left( \frac{\exp(\text{score}(e_j))}{\sum_{k=1}^{|\mathcal{E}|} \exp(\text{score}(e_k))} \right),
\end{equation}
where $y_j^{\text{LS}}$ are the smoothed one-hot labels for the ground-truth entity.
The final training objective $\mathcal{L}$ is a weighted sum of this prediction loss, the diffusion loss $\mathcal{L}_{\text{diff}}$ (Section~\ref{subsec:diffusion}), and an auxiliary timestamp prediction loss $\mathcal{L}_{\text{time}}$ (Section~\ref{subsec:transformer}):
\begin{equation}
\mathcal{L} = \mathcal{L}_{\text{pred}} + \lambda_{\text{diff}}\mathcal{L}_{\text{diff}} + \lambda_{\text{time}}\mathcal{L}_{\text{time}},
\end{equation}
where $\lambda_{\text{diff}}$ and $\lambda_{\text{time}}$ are hyperparameters that balance the contribution of each auxiliary task. This multi-task learning setup encourages the model to learn more generalizable representations.

\section{Experiments}
\begin{table*}[h!]
  \caption{Performance of extrapolation reasoning task on GDELT, ICEWS05-15, and ICEWS18. The best performances are highlighted in \textbf{bold}, and the second-best are \underline{underlined}. The symbol "-" indicates that the corresponding baseline did not report results for that specific metric or dataset.}
  \label{ex_exam}
  \centering
  \small
  \setlength{\tabcolsep}{4pt} 
  \resizebox{\linewidth}{!}{%
  \begin{tabular}{lcccccccccccc}
    \toprule
    & \multicolumn{4}{c}{\textbf{GDELT}} & \multicolumn{4}{c}{\textbf{ICEWS05-15}} & \multicolumn{4}{c}{\textbf{ICEWS18}} \\
    \cmidrule(r){2-5} \cmidrule(r){6-9} \cmidrule(r){10-13}
    Approach & MRR & H@1 & H@3 & H@10 & MRR & H@1 & H@3 & H@10 & MRR & H@1 & H@3 & H@10 \\
    \midrule
    RE-GCN & 19.69  & 12.46  & 20.93  & 33.81  & 48.03  & 37.33  & 53.90  & 68.51  & 32.62  & 22.39  & 36.79  & 52.68  \\
    L2TKG & 20.53  & 12.89  & -     & 35.83  & 57.43  & 41.86  & -     & 80.69  & 33.36  & 22.15  & -     & 55.04  \\
    DREAM & 28.10  & 19.30  & 31.10  & 44.70  & 56.80  & 47.30  & 65.10  & 78.60  & 39.10  & 28.00  & \textbf{45.20}  & \textbf{62.70}  \\
    LogCL & 23.75  & 14.64  & 25.60  & 42.33  & 57.04  & 46.07  & 63.72  & 77.87  & 35.67  & 24.53  & 40.32  & \underline{57.74}  \\
    DyMemR & 25.46  & 16.79  & 27.96  & 42.49  & 53.76  & 44.68  & 58.66  & 70.84  & 35.50  & 25.29  & 39.90  & 55.44  \\
    THCN  & 23.46  & 15.18  & 25.21  & 39.03  & 51.94  & 40.32  & 57.79  & 72.18  & 35.63  & 24.90  & 39.26  & 56.76  \\
    \midrule
    CSI   & 25.60  & 18.20  & 28.50  & 40.70  & 53.30  & 42.50  & 59.20  & 73.60  & -     & -     & -     & - \\
    TempValid & 21.88  & 14.37  & 24.40  & 37.00  & 50.31  & 39.46  & 56.71  & 70.55  & 33.50  & 23.91  & 37.89  & 52.33  \\
    LLMDA & -     & -     & -     & -     & 52.10  & 41.60  & 58.60  & 72.80  & 35.70  & 25.50  & 40.30  & 57.00  \\
    \midrule
    TPAR & -     & -     & -     & -     & -     & -     & -     & -     & 35.76  & 26.58  & 39.27  & 56.94  \\
    ECEformer & \underline{50.89}  & \underline{38.71}  & \underline{59.45}  & \underline{71.02}  & \underline{68.47}  & \underline{61.59}  & \underline{72.29}  & \underline{81.22}  & \underline{40.13}  & \underline{32.25}  & 43.63  & 54.63  \\ 
    \midrule
    \textbf{DynaGen}  & \textbf{52.41} & \textbf{39.99} & \textbf{60.59} & \textbf{72.22} & \textbf{69.84} & \textbf{63.52} & \textbf{73.10} & \textbf{81.85} & \textbf{41.58}  & \textbf{33.80}  & \underline{45.04}  & 56.06  \\
    \bottomrule
  \end{tabular}}
\end{table*}

\begin{table*}[h!]
  \caption{Performance of interpolation reasoning task on ICEWS14, YAGO, and Wikidata. The best performances are highlighted in \textbf{bold}, and the second-best are \underline{underlined}. The symbol "-" indicates that the corresponding baseline did not report results for that specific metric or dataset.}
  \label{inter_exam}
  \small
  \centering
  
  \setlength{\tabcolsep}{4pt}
  \resizebox{\linewidth}{!}{%
  \begin{tabular}{lcccccccccccc}
    \toprule
    & \multicolumn{4}{c}{\textbf{ICEWS14}} & \multicolumn{4}{c}{\textbf{Yago}} & \multicolumn{4}{c}{\textbf{Wikidata}} \\
    \cmidrule(r){2-5} \cmidrule(r){6-9} \cmidrule(r){10-13}
    Approach & MRR & H@1 & H@3 & H@10 & MRR & H@1 & H@3 & H@10 & MRR & H@1 & H@3 & H@10 \\
    \midrule
    RotateQVS & 59.10 & 50.70 & 64.20 & 75.40 & 18.90 & 12.40 & 19.90 & 32.30 & - & - & - & - \\
    TeAST & 63.70 & 56.00 & 68.20 & 78.20 & - & - & - & - & - & - & - & - \\
    HYIE & 63.10 & 56.30 & 68.70 & 78.60 & 19.10 & 12.50 & 20.10 & 32.60 & 30.10 & 19.70 & 32.80 & 50.60 \\
    LCGE & 66.70 & 58.80 & 71.40 & \textbf{81.50} & - & - & - & - & 42.90 & 30.40 & \underline{49.50} & \textbf{67.70} \\
    HGE & 63.40 & 55.00 & 68.50 & 78.80 & - & - & - & - & - & - & - & - \\
    5EL & 64.20 & 57.10 & 69.50 & 80.10 & 21.70 & 18.10 & 25.80 & \underline{39.70} & 31.10 & 23.70 & 35.50 & 54.60 \\
    TGeomE & 62.90 & 54.60 & 68.00 & 78.00 & 19.50 & 13.00 & 19.60 & 32.60 & 33.30 & 23.20 & 36.20 & 54.60 \\
    \midrule
    TPAR & 65.07 & 57.03 & 69.74 & 80.41 & 24.12 & 17.35 & 25.14 & 37.42 & 34.89 & 25.05 & 38.68 & 54.79 \\
    ECEformer & \underline{69.46} & \underline{64.80} & \underline{71.81} & 78.14 & \underline{25.13} & \underline{19.14} & \underline{26.34} & 37.21 & \underline{45.24} & \underline{38.50} & 47.92 & 58.03 \\ 
    \midrule
    \textbf{DynaGen} & \textbf{72.14} & \textbf{67.97} & \textbf{77.72} & \underline{80.52} & \textbf{27.83} & \textbf{21.46} & \textbf{29.38} & \textbf{40.34} & \textbf{47.69} & \textbf{40.59} & \textbf{49.67} & \underline{60.24} \\
    \bottomrule
  \end{tabular}}
\end{table*}

\subsection{Experimental Settings}
\label{Experimental Settings}

\paragraph{Datasets.} We evaluate the performance of \textbf{DynaGen} on six Temporal Knowledge Graph (TKG) benchmark datasets: ICEWS05-15 \cite{boschee2015icews}, ICEWS18 \cite{boschee2015icews}, ICEWS14 \cite{boschee2015icews}, GDELT \cite{leetaru13gdelt}, YAGO \cite{suchanek2007yago}, WIKI \cite{vrandevcic2014wikidata}. For interpolation reasoning, we split the dataset of GDELT and ICEWS05-15/18 by timestamps~\cite{liang2023rpc}. And for extrapolation reasoning, we split the dataset of ICEWS14, YAGO, and WIKI by time intervals~\cite{dasgupta2018hyte}. Table~\ref{Table: dataset} presents the statistics of the six temporal knowledge graph datasets used in our experiments.

\paragraph{Experimental Settings.}Our model is implemented in PyTorch and trained on a single NVIDIA RTX 3090 GPU. We set the embedding dimension to 320. The core architecture consists of a 2-layer SSAE for subgraph encoding and an 8-layer Transformer for the reasoning module. For training, we use the Adamax optimizer with a learning rate of 0.02. A complete list of hyperparameters is provided in Appendix~\ref{sec:full_experimental_settings}.

\paragraph{Baselines.} We evaluate our proposed DynaGen method against 18 state-of-the-art (SOTA) Temporal Knowledge Graph Reasoning (TKGR) approaches. For extrapolation tasks, we compare against three categories of baselines: (1) GNN-based methods including RE-GCN \cite{li2021regcn}, L2TKG \cite{zhang2023l2tkg}, DREAM \cite{zheng2023dream}, LogCL \cite{chen2024logcl}, DyMemR \cite{zhang2024dymemr}, and THCN \cite{chen2024thcn}; (2) other specialized methods such as TempValid \cite{huang2024tempvalid}, CSI \cite{chen2024csi}, and LLMDA \cite{wang2024llmda}; and (3) unified methods like TPAR \cite{chen2024unified} and ECEformer \cite{fang2024eceformer}. For interpolation tasks, we benchmark against geometric methods—RotateQVS \cite{chen2022rotateqvs}, TeAST \cite{li2023teast}, HYIE \cite{zhang2023hyie}, LCGE \cite{Niu2023lcge}, HGE \cite{pan2024hge}, 5EL \cite{zhang20255el}, and TGeomE \cite{xu2023tgeome}—as well as the unified methods TPAR \cite{chen2024unified} and ECEformer \cite{fang2024eceformer}.

\subsection{Performance Comparison}
\label{Performance Comparison}

As shown in Tables \ref{ex_exam} and \ref{inter_exam}, DynaGen achieves state-of-the-art performance on all six datasets across both tasks, demonstrating its capability in capturing temporal dynamics and performing accurate reasoning over temporal knowledge graphs. 

Specifically, in extrapolation reasoning (Table \ref{ex_exam}), DynaGen outperforms all other models on GDELT, ICEWS05-15, and ICEWS18, which improves over the second-best models by 1.52, 1.37, and 1.45 points in MRR. In interpolation reasoning (Table \ref{inter_exam}), DynaGen consistently delivers the best performance across all datasets (ICEWS14, Yago, and Wikidata), which improves over the second-best models by 2.68, 2.70, and 2.45 points in MRR.

These results highlight DynaGen's effectiveness across a wide range of datasets and tasks. The consistent superiority of DynaGen underscores its ability to handle both extrapolation and interpolation reasoning tasks with high accuracy.

\subsection{Ablation study}
\label{Ablation study}

\begin{table}[htbp]
  \centering
  \caption{The performance of DynaGen and the variants. The best results are in \textbf{bold}.}
  \label{ab_table}
  \small
  \setlength{\tabcolsep}{7pt}
  \renewcommand{\arraystretch}{1}
  \resizebox{\linewidth}{!}{
  \begin{tabular}{lcccc}
    \toprule
    & \multicolumn{4}{c}{\textbf{Yago}}   \\  
    \cmidrule(r){2-5}
    Approach & MRR   & Hit@1 & Hit@3 & Hit@10 \\
    \midrule
    w/o SSAE & 25.75  & 19.22  & 26.83  & 37.96  \\
    w/o Diffusion & 26.14  & 19.91  & 27.32  & 38.44  \\
    DynaGen  & \textbf{27.83} & \textbf{21.46} & \textbf{29.38} & \textbf{40.34} \\
    \midrule
    & \multicolumn{4}{c}{\textbf{ICEWS05-15}} \\
    \cmidrule(r){2-5}
    Approach & MRR & Hit@1 & Hit@3 & Hit@10 \\
    \midrule
    w/o SSAE & 68.34  & 61.75  & 72.45  & 81.23 \\
    w/o Diffusion & 67.85  & 61.05  & 71.87  & 80.54  \\
    DynaGen   & \textbf{69.84} & \textbf{63.52} & \textbf{73.10} & \textbf{81.85} \\
    \bottomrule
  \end{tabular}}
\end{table}

As shown in Table \ref{ab_table}, we draw several conclusions from the ablation study:
\textbf{(1)} The most substantial performance degradation is observed in the DynaGen-w/o SSAE variant, which demonstrates the critical importance of the SSAE component in capturing complex temporal patterns. For instance, on Yago, removing SSAE leads to a 2.08 point drop in MRR and a 2.38 point drop in Hits@10. \textbf{(2)} The performance degradation of the DynaGen-w/o diffusion variant, though less severe, still validates the effectiveness of the diffusion mechanism in enhancing the model's predictive capability. \textbf{(3)} The consistent performance improvements across both datasets confirm that each component contributes meaningfully to the overall model effectiveness.

\subsection{Influence of SSAE Layer Depth}
\label{ex_layer}

\begin{figure}[htbp]
  \centering
  \includegraphics[width=\linewidth]{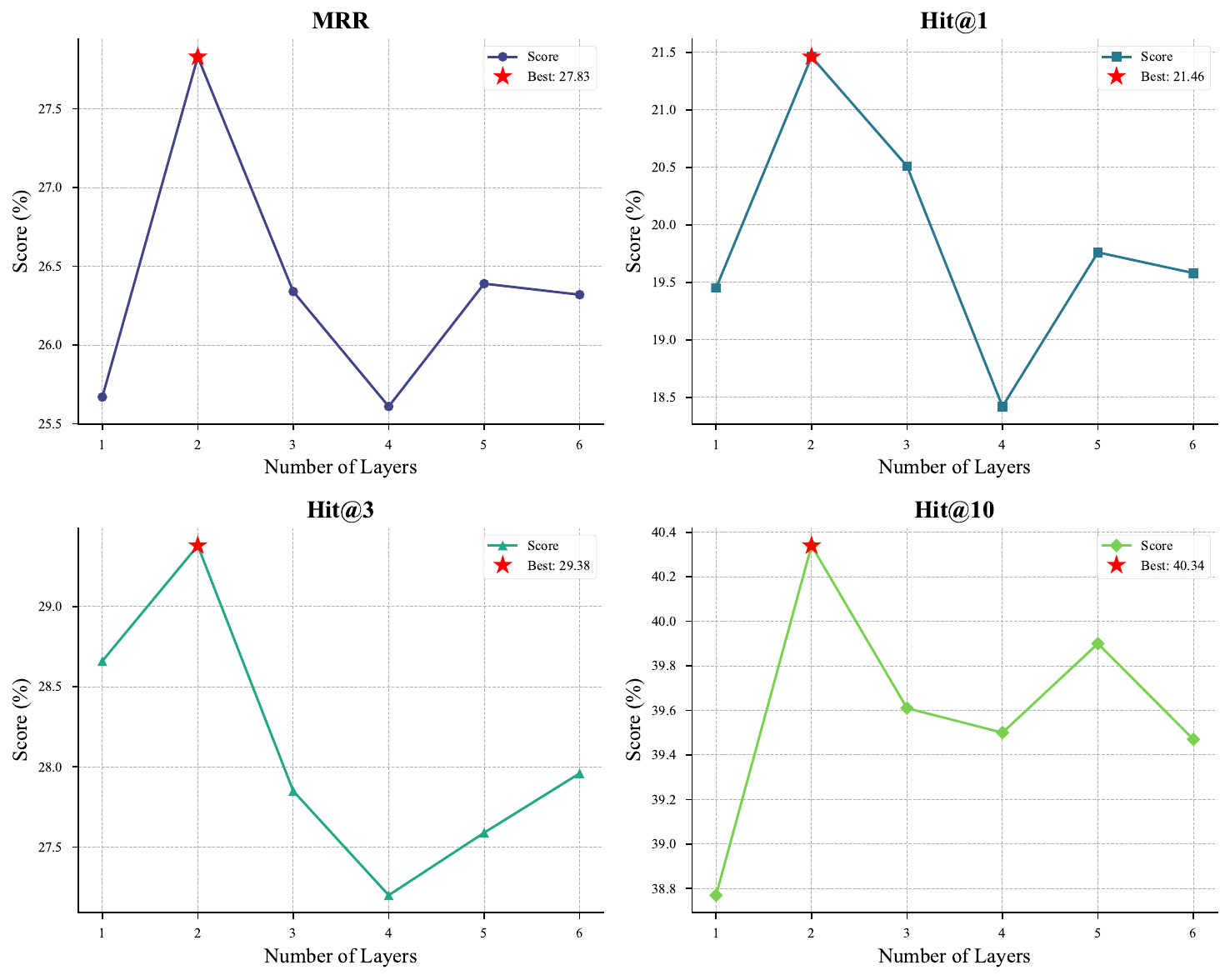}
  \caption{Variation of key link-prediction metrics (MRR, Hit@1, Hit@3, Hit@10) as a function of the number of SSAE layers.}
  \label{layers}
\end{figure}

As shown in Figure~\ref{layers}, model performance improves significantly as we increase the model depth from 1 to 2 layers. Specifically, MRR increases from 25.67 to \textbf{27.83}, Hit@1 from 19.45 to \textbf{21.46}, Hit@3 from 28.66 to \textbf{29.38}, and Hit@10 from 38.77 to \textbf{40.34}. All metrics reach their peak at 2 layers.
However, after exceeding two layers, the model's overall performance begins to decline. This performance degradation suggests that deeper architectures may lead to issues such as over-smoothing. Although some metrics exhibit minor fluctuations, they never surpass the optimal performance achieved by the 2-layer model.
In summary, the results demonstrate that a two-layer model architecture achieves the best overall performance in this evaluation.

\section{Conclusion}

In this paper, we introduced DynaGen, a unified framework for temporal knowledge graph reasoning. DynaGen addresses the distinct challenges of interpolation and extrapolation through two core innovations: (i) A synergistic dual-branch GNN processes dynamic, entity-centric subgraphs to capture the rich structural context required for accurate interpolation. (ii) A novel generative regularization strategy compelling the model to learn the underlying principles of temporal evolution for robust extrapolation. Comprehensive experiments on six benchmark datasets demonstrate that DynaGen achieves state-of-the-art performance. 

\section*{Limitations}
Our approach improves the performance of TKGR, but there are still some possible limitations. First, the heuristic-based method for subgraph extraction may still incorporate noisy connections or overlook temporally distant yet logically crucial facts. Exploring learned, rather than rule-based, subgraph generation methods presents a promising future direction to create more accurate and relevant context for reasoning. Second, the process of dynamically building and encoding a unique subgraph for every query during training is resource-intensive, which introduces significant computational overhead. Future work could investigate more efficient indexing structures or graph sampling techniques to reduce this overhead.

\section*{Ethics Statement}
This study employed six publicly available temporal knowledge graph benchmark datasets (ICEWS05-15, ICEWS18, ICEWS14, GDELT, YAGO, and WIKI) in strict accordance with their respective licenses and usage guidelines.   Nevertheless, our work entails important ethical considerations.   First, while the DynaGen model demonstrates strong performance on temporal-reasoning tasks, it may inherit biases present in the training data.   By learning historical patterns embedded in the datasets, the model’s predictions can disproportionately reflect those biases.   We emphasize that this work focuses primarily on technical improvements in temporal knowledge graph reasoning;   any real-world deployment must therefore be accompanied by dedicated bias-detection and mitigation strategies. Second, Predictions generated by the model should be treated as probabilistic inferences derived from historical data rather than established facts.   The outputs are directly contingent on the quality and coverage of the training data, which may themselves contain errors, outdated information, or omissions.    In practical applications, we strongly recommend appending confidence metrics to all model outputs and instituting rigorous human-review protocols to prevent misinterpretation or misuse of uncertain information.

\bibliography{custom}

@article{boschee2015icews,
author = {Boschee, Elizabeth and Lautenschlager, Jennifer and O'Brien, Sean and Shellman, Steve and Starz, James and Ward, Michael},
publisher = {Harvard Dataverse},
title = {{ICEWS Coded Event Data}},
UNF = {UNF:6:NOSHB7wyt0SQ8sMg7+w38w==},
year = {2015},
version = {V37},
doi = {10.7910/DVN/28075},
url = {https://doi.org/10.7910/DVN/28075}
}

@article{leetaru13gdelt,
  title={Gdelt: Global data on events, location, and tone, 1979--2012},
  author={Leetaru, Kalev and Schrodt, Philip A},
  booktitle={ISA annual convention},
  volume={2},
  pages={1--49},
  year={2013},
  organization={Citeseer}
}

@article{suchanek2007yago,
  title={Yago: a core of semantic knowledge},
  author={Suchanek, Fabian M and Kasneci, Gjergji and Weikum, Gerhard},
  booktitle={Proceedings of the 16th international conference on World Wide Web},
  pages={697--706},
  year={2007}
}

@article{vrandevcic2014wikidata,
  title={Wikidata: a free collaborative knowledgebase},
  author={Vrande{\v{c}}i{\'c}, Denny and Kr{\"o}tzsch, Markus},
  journal={Communications of the ACM},
  volume={57},
  number={10},
  pages={78--85},
  year={2014},
  publisher={ACM New York, NY, USA}
}

@article{fang2024eceformer,
  title={Transformer-based reasoning for learning evolutionary chain of events on temporal knowledge graph},
  author={Fang, Zhiyu and Lei, Shuai-Long and Zhu, Xiaobin and Yang, Chun and Zhang, Shi-Xue and Yin, Xu-Cheng and Qin, Jingyan},
  booktitle={Proceedings of the 47th International ACM SIGIR Conference on Research and Development in Information Retrieval},
  pages={70--79},
  year={2024}
}

@article{wang2024llmda,
  title={Large language models-guided dynamic adaptation for temporal knowledge graph reasoning},
  author={Wang, Jiapu and Kai, Sun and Luo, Linhao and Wei, Wei and Hu, Yongli and Liew, Alan Wee-Chung and Pan, Shirui and Yin, Baocai},
  journal={Advances in Neural Information Processing Systems},
  volume={37},
  pages={8384--8410},
  year={2024}
}

@article{chen2024decrl,
  title={DECRL: A Deep Evolutionary Clustering Jointed Temporal Knowledge Graph Representation Learning Approach},
  author={Chen, Qian and Chen, Ling},
  journal={Advances in Neural Information Processing Systems},
  volume={37},
  pages={55204--55227},
  year={2024}
}

@article{lin2023tflex,
  title={TFLEX: Temporal feature-logic embedding framework for complex reasoning over temporal knowledge graph},
  author={Lin, Xueyuan and Xu, Chengjin and Zhou, Gengxian and Luo, Haoran and Hu, Tianyi and Su, Fenglong and Li, Ningyuan and Sun, Mingzhi and others},
  journal={Advances in Neural Information Processing Systems},
  volume={36},
  pages={73039--73081},
  year={2023}
}

@article{liang2023rpc,
  title={Learn from relational correlations and periodic events for temporal knowledge graph reasoning},
  author={Liang, Ke and Meng, Lingyuan and Liu, Meng and Liu, Yue and Tu, Wenxuan and Wang, Siwei and Zhou, Sihang and Liu, Xinwang},
  booktitle={Proceedings of the 46th international ACM SIGIR conference on research and development in information retrieval},
  pages={1559--1568},
  year={2023}
}

@article{dasgupta2018hyte,
  title={Hyte: Hyperplane-based temporally aware knowledge graph embedding},
  author={Dasgupta, Shib Sankar and Ray, Swayambhu Nath and Talukdar, Partha},
  booktitle={Proceedings of the 2018 conference on empirical methods in natural language processing},
  pages={2001--2011},
  year={2018}
}

@article{chen2024unified,
    title = "A Unified Temporal Knowledge Graph Reasoning Model Towards Interpolation and Extrapolation",
    author = "Chen, Kai  and
      Wang, Ye  and
      Li, Yitong  and
      Li, Aiping  and
      Yu, Han  and
      Song, Xin",
    editor = "Ku, Lun-Wei  and
      Martins, Andre  and
      Srikumar, Vivek",
    booktitle = "Proceedings of the 62nd Annual Meeting of the Association for Computational Linguistics (Volume 1: Long Papers)",
    month = aug,
    year = "2024",
    address = "Bangkok, Thailand",
    publisher = "Association for Computational Linguistics",
    url = "https://aclanthology.org/2024.acl-long.8/",
    doi = "10.18653/v1/2024.acl-long.8",
    pages = "117--132"
}

@article{cao2025dpcl_diff,
  title={DPCL-Diff: Temporal Knowledge Graph Reasoning Based on Graph Node Diffusion Model with Dual-Domain Periodic Contrastive Learning},
  author={Cao, Yukun and Wang, Lisheng and Huang, Luobin},
  booktitle={Proceedings of the AAAI Conference on Artificial Intelligence},
  volume={39},
  pages={14806--14814},
  year={2025}
}

@article{zhang20255el,
  title={Integrating Large Language Models and M{\"o}bius Group Transformations for Temporal Knowledge Graph Embedding on the Riemann Sphere},
  author={Zhang, Sensen and Liang, Xun and Niu, Simin and Niu, Zhendong and Wu, Bo and Hua, Gengxin and Wang, Long and Guan, Zhenyu and Wang, Hanyu and Zhang, Xuan and others},
  booktitle={Proceedings of the AAAI Conference on Artificial Intelligence},
  volume={39},
  pages={13277--13285},
  year={2025}
}

@article{pan2024hge,
  title={HGE: embedding temporal knowledge graphs in a product space of heterogeneous geometric subspaces},
  author={Pan, Jiaxin and Nayyeri, Mojtaba and Li, Yinan and Staab, Steffen},
  booktitle={Proceedings of the AAAI Conference on Artificial Intelligence},
  volume={38},
  pages={8913--8920},
  year={2024}
}

@article{huang2024tempvalid,
  title={Confidence is not timeless: Modeling temporal validity for rule-based temporal knowledge graph forecasting},
  author={Huang, Rikui and Wei, Wei and Qu, Xiaoye and Zhang, Shengzhe and Chen, Dangyang and Cheng, Yu},
  booktitle={Proceedings of the 62nd Annual Meeting of the Association for Computational Linguistics (Volume 1: Long Papers)},
  pages={10783--10794},
  year={2024}
}

@article{ying2024tcompounde,
    title = "Simple but Effective Compound Geometric Operations for Temporal Knowledge Graph Completion",
    author = "Ying, Rui  and
      Hu, Mengting  and
      Wu, Jianfeng  and
      Xie, Yalan  and
      Liu, Xiaoyi  and
      Wang, Zhunheng  and
      Jiang, Ming  and
      Gao, Hang  and
      Zhang, Linlin  and
      Cheng, Renhong",
    editor = "Ku, Lun-Wei  and
      Martins, Andre  and
      Srikumar, Vivek",
    booktitle = "Proceedings of the 62nd Annual Meeting of the Association for Computational Linguistics (Volume 1: Long Papers)",
    month = aug,
    year = "2024",
    address = "Bangkok, Thailand",
    publisher = "Association for Computational Linguistics",
    url = "https://aclanthology.org/2024.acl-long.596/",
    doi = "10.18653/v1/2024.acl-long.596",
    pages = "11074--11086"
}

@article{xin2024kartgps,
  title={KartGPS: Knowledge Base Update with Temporal Graph Pattern-based Semantic Rules},
  author={Xin, Hao and Chen, Lei},
  booktitle={2024 IEEE 40th International Conference on Data Engineering (ICDE)},
  pages={5075--5087},
  year={2024},
  organization={IEEE}
}

@article{chen2024logcl,
  title={Local-global history-aware contrastive learning for temporal knowledge graph reasoning},
  author={Chen, Wei and Wan, Huaiyu and Wu, Yuting and Zhao, Shuyuan and Cheng, Jiayaqi and Li, Yuxin and Lin, Youfang},
  booktitle={2024 IEEE 40th International Conference on Data Engineering (ICDE)},
  pages={733--746},
  year={2024},
  organization={IEEE}
}

@article{chen2024csi,
  title={Temporal knowledge graph extrapolation via causal subhistory identification},
  author={Chen, Kai and Wang, Ye and Song, Xin and Chen, Siwei and Yu, Han and Li, Aiping},
  booktitle={Proceedings of the Thirty-Third International Joint Conference on Artificial Intelligence},
  pages={3298--3306},
  year={2024}
}

@article{tang2024gtrl,
  title={GTRL: An entity group-aware temporal knowledge graph representation learning method},
  author={Tang, Xing and Chen, Ling},
  journal={IEEE Transactions on Knowledge and Data Engineering},
  volume={36},
  number={9},
  pages={4707--4721},
  year={2023},
  publisher={IEEE}
}

@article{zhang2024dymemr,
  title={Temporal Knowledge Graph Reasoning With Dynamic Memory Enhancement},
  author={Zhang, Fuwei and Zhang, Zhao and Zhuang, Fuzhen and Zhao, Yu and Wang, Deqing and Zheng, Hongwei},
  journal={IEEE Transactions on Knowledge and Data Engineering},
  year={2024},
  publisher={IEEE}
}

@article{chen2024thcn,
  title={THCN: A Hawkes Process Based Temporal Causal Convolutional Network for Extrapolation Reasoning in Temporal Knowledge Graphs},
  author={Chen, Tingxuan and Long, Jun and Wang, Zidong and Luo, Shuai and Huang, Jincai and Yang, Liu},
  journal={IEEE Transactions on Knowledge and Data Engineering},
  year={2024},
  publisher={IEEE}
}

@article{tang2024dhyper,
  title={Dhyper: a recurrent dual hypergraph neural network for event prediction in temporal knowledge graphs},
  author={Tang, Xing and Chen, Ling and Shi, Hongyu and Lyu, Dandan},
  journal={ACM Transactions on Information Systems},
  volume={42},
  number={5},
  pages={1--23},
  year={2024},
  publisher={ACM New York, NY}
}

@article{zheng2023dream,
  title={DREAM: Adaptive reinforcement learning based on attention mechanism for temporal knowledge graph reasoning},
  author={Zheng, Shangfei and Yin, Hongzhi and Chen, Tong and Nguyen, Quoc Viet Hung and Chen, Wei and Zhao, Lei},
  booktitle={Proceedings of the 46th international ACM SIGIR conference on research and development in information retrieval},
  pages={1578--1588},
  year={2023}
}

@article{xu2023tgeome,
  title={Geometric algebra based embeddings for static and temporal knowledge graph completion},
  author={Xu, Chengjin and Nayyeri, Mojtaba and Chen, Yung-Yu and Lehmann, Jens},
  journal={IEEE Transactions on Knowledge and Data Engineering},
  volume={35},
  number={5},
  pages={4838--4851},
  year={2022},
  publisher={IEEE}
}

@article{zhang2023l2tkg,
  title={Learning latent relations for temporal knowledge graph reasoning},
  author={Zhang, Mengqi and Xia, Yuwei and Liu, Qiang and Wu, Shu and Wang, Liang},
  booktitle={Proceedings of the 61st Annual Meeting of the Association for Computational Linguistics (Volume 1: Long Papers)},
  pages={12617--12631},
  year={2023}
}

@article{Niu2023lcge, 
  title={Logic and commonsense-guided temporal knowledge graph completion},
  author={Niu, Guanglin and Li, Bo},
  booktitle={Proceedings of the AAAI Conference on Artificial Intelligence},
  volume={37},
  number={4},
  pages={4569--4577},
  journal={Proceedings of the AAAI Conference on Artificial Intelligence},
  year={2023}
}

@article{chen2023mtkge,
  title     = {Meta-Learning Based Knowledge Extrapolation for Knowledge Graphs in the Federated Setting},
  author    = {Chen, Mingyang and Zhang, Wen and Yao, Zhen and Chen, Xiangnan and Ding, Mengxiao and Huang, Fei and Chen, Huajun},
  booktitle = {Proceedings of the Thirty-First International Joint Conference on
               Artificial Intelligence, {IJCAI-22}},
  publisher = {International Joint Conferences on Artificial Intelligence Organization},
  editor    = {Lud De Raedt},
  pages     = {1966--1972},
  year      = {2022},
  month     = {7},
  note      = {Main Track},
  doi       = {10.24963/ijcai.2022/273},
  url       = {https://doi.org/10.24963/ijcai.2022/273},
}

@article{liu2023retia,
  author={Liu, Kangzheng and Zhao, Feng and Xu, Guandong and Wang, Xianzhi and Jin, Hai},
  booktitle={2023 IEEE 39th International Conference on Data Engineering (ICDE)}, 
  title={RETIA: Relation-Entity Twin-Interact Aggregation for Temporal Knowledge Graph Extrapolation}, 
  year={2023},
  volume={},
  number={},
  pages={1761-1774},
  doi={10.1109/ICDE55515.2023.00138}
}

@article{li2023teast,
    title = "{T}e{AST}: Temporal Knowledge Graph Embedding via Archimedean Spiral Timeline",
    author = "Li, Jiang  and
      Su, Xiangdong  and
      Gao, Guanglai",
    editor = "Rogers, Anna  and
      Boyd-Graber, Jordan  and
      Okazaki, Naoaki",
    booktitle = "Proceedings of the 61st Annual Meeting of the Association for Computational Linguistics (Volume 1: Long Papers)",
    month = jul,
    year = "2023",
    address = "Toronto, Canada",
    publisher = "Association for Computational Linguistics",
    url = "https://aclanthology.org/2023.acl-long.862/",
    doi = "10.18653/v1/2023.acl-long.862",
    pages = "15460--15474"
}

@article{xu2023cenet, 
 title={Temporal Knowledge Graph Reasoning with Historical Contrastive Learning}, 
 volume={37}, 
 url={https://ojs.aaai.org/index.php/AAAI/article/view/25601}, 
 DOI={10.1609/aaai.v37i4.25601},
 number={4}, 
 journal={Proceedings of the AAAI Conference on Artificial Intelligence}, 
 author={Xu, Yi and Ou, Junjie and Xu, Hui and Fu, Luoyi}, 
 year={2023}, 
 month={Jun.}, 
 pages={4765-4773} 
}

@article{schlichtkrull2018rgcn,
author = {Schlichtkrull, Michael and Kipf, Thomas N. and Bloem, Peter and van\&nbsp;den Berg, Rianne and Titov, Ivan and Welling, Max},
title = {Modeling Relational Data with Graph Convolutional Networks},
year = {2018},
isbn = {978-3-319-93416-7},
publisher = {Springer-Verlag},
address = {Berlin, Heidelberg},
url = {https://doi.org/10.1007/978-3-319-93417-4_38},
doi = {10.1007/978-3-319-93417-4_38},
booktitle = {The Semantic Web: 15th International Conference, ESWC 2018, Heraklion, Crete, Greece, June 3–7, 2018, Proceedings},
pages = {593–607},
numpages = {15},
location = {Heraklion, Greece}
}

@article{jin2020renet,
    title = "Recurrent Event Network: Autoregressive Structure Inferenceover Temporal Knowledge Graphs",
    author = "Jin, Woojeong  and
      Qu, Meng  and
      Jin, Xisen  and
      Ren, Xiang",
    editor = "Webber, Bonnie  and
      Cohn, Trevor  and
      He, Yulan  and
      Liu, Yang",
    booktitle = "Proceedings of the 2020 Conference on Empirical Methods in Natural Language Processing (EMNLP)",
    month = nov,
    year = "2020",
    address = "Online",
    publisher = "Association for Computational Linguistics",
    url = "https://aclanthology.org/2020.emnlp-main.541/",
    doi = "10.18653/v1/2020.emnlp-main.541",
    pages = "6669--6683"
}

@article{leblay2018deriving,
author = {Leblay, Julien and Chekol, Melisachew Wudage},
title = {Deriving Validity Time in Knowledge Graph},
year = {2018},
isbn = {9781450356404},
publisher = {International World Wide Web Conferences Steering Committee},
address = {Republic and Canton of Geneva, CHE},
url = {https://doi.org/10.1145/3184558.3191639},
doi = {10.1145/3184558.3191639},
booktitle = {Companion Proceedings of the The Web Conference 2018},
pages = {1771–1776},
numpages = {6},
location = {Lyon, France},
series = {WWW '18}
}

@article{messner2022boxte, 
 title={Temporal Knowledge Graph Completion Using Box Embeddings}, 
 volume={36}, 
 url={https://ojs.aaai.org/index.php/AAAI/article/view/20746}, 
 DOI={10.1609/aaai.v36i7.20746}, 
 abstractNote={Knowledge graph completion is the task of inferring missing facts based on existing data in a knowledge graph. Temporal knowledge graph completion (TKGC) is an extension of this task to temporal knowledge graphs, where each fact is additionally associated with a time stamp. Current approaches for TKGC primarily build on existing embedding models which are developed for static knowledge graph completion, and extend these models to incorporate time, where the idea is to learn latent representations for entities, relations, and timestamps and then use the learned representations to predict missing facts at various time steps. In this paper, we propose BoxTE, a box embedding model for TKGC, building on the static knowledge graph embedding model BoxE. We show that BoxTE is fully expressive, and possesses strong inductive capacity in the temporal setting. We then empirically evaluate our model and show that it achieves state-of-the-art results on several TKGC benchmarks}, 
 number={7}, 
 journal={Proceedings of the AAAI Conference on Artificial Intelligence}, 
 author={Messner, Johannes and Abboud, Ralph and Ceylan, Ismail Ilkan}, 
 year={2022}, 
 month={Jun.}, 
 pages={7779-7787} 
}

@article{chen2022rotateqvs,
    title = "{R}otate{QVS}: Representing Temporal Information as Rotations in Quaternion Vector Space for Temporal Knowledge Graph Completion",
    author = "Chen, Kai  and
      Wang, Ye  and
      Li, Yitong  and
      Li, Aiping",
    editor = "Muresan, Smaranda  and
      Nakov, Preslav  and
      Villavicencio, Aline",
    booktitle = "Proceedings of the 60th Annual Meeting of the Association for Computational Linguistics (Volume 1: Long Papers)",
    month = may,
    year = "2022",
    address = "Dublin, Ireland",
    publisher = "Association for Computational Linguistics",
    url = "https://aclanthology.org/2022.acl-long.402/",
    doi = "10.18653/v1/2022.acl-long.402",
    pages = "5843--5857"
}

@article{zhang2023hyie,
author = {Zhang, Sensen and Liang, Xun and Tang, Hui and Guan, Zhenyu},
title = {Hybrid Interaction Temporal Knowledge Graph Embedding Based on Householder Transformations},
year = {2023},
isbn = {9798400701085},
publisher = {Association for Computing Machinery},
address = {New York, NY, USA},
url = {https://doi.org/10.1145/3581783.3613446},
doi = {10.1145/3581783.3613446},
booktitle = {Proceedings of the 31st ACM International Conference on Multimedia},
pages = {8954–8962},
numpages = {9},
location = {Ottawa ON, Canada},
series = {MM '23}
}

@article{wang2019kgat,
  title={Kgat: Knowledge graph attention network for recommendation},
  author={Wang, Xiang and He, Xiangnan and Cao, Yixin and Liu, Meng and Chua, Tat-Seng},
  booktitle={Proceedings of the 25th ACM SIGKDD international conference on knowledge discovery \& data mining},
  pages={950--958},
  year={2019}
}

@article{sang2022dialoguegat,
  title={DialogueGAT: A graph attention network for financial risk prediction by modeling the dialogues in earnings conference calls},
  author={Sang, Yunxin and Bao, Yang},
  booktitle={Findings of the Association for Computational Linguistics: EMNLP 2022},
  pages={1623--1633},
  year={2022}
}

@article{liu2018entity,
    title = "Entity-Duet Neural Ranking: Understanding the Role of Knowledge Graph Semantics in Neural Information Retrieval",
    author = "Liu, Zhenghao  and
      Xiong, Chenyan  and
      Sun, Maosong  and
      Liu, Zhiyuan",
    editor = "Gurevych, Iryna  and
      Miyao, Yusuke",
    booktitle = "Proceedings of the 56th Annual Meeting of the Association for Computational Linguistics (Volume 1: Long Papers)",
    month = jul,
    year = "2018",
    address = "Melbourne, Australia",
    publisher = "Association for Computational Linguistics",
    url = "https://aclanthology.org/P18-1223/",
    doi = "10.18653/v1/P18-1223",
    pages = "2395--2405"
}

@article{li2021regcn,
  title={Temporal knowledge graph reasoning based on evolutional representation learning},
  author={Li, Zixuan and Jin, Xiaolong and Li, Wei and Guan, Saiping and Guo, Jiafeng and Shen, Huawei and Wang, Yuanzhuo and Cheng, Xueqi},
  booktitle={Proceedings of the 44th international ACM SIGIR conference on research and development in information retrieval},
  pages={408--417},
  year={2021}
}

\clearpage

\appendix
This appendix provides supplementary material to ensure the clarity, depth, and reproducibility of our work. It is structured as follows: Section A offers a comprehensive overview of our experimental framework, including detailed statistics of the datasets, a full list of experimental settings and hyperparameters, descriptions of the baseline models against which DynaGen was compared, and qualitative case studies for both interpolation and extrapolation tasks. Section B delves into finer methodological details, providing an in-depth explanation of the temporal subgraph construction algorithm, the specific architecture of the Synergistic Structure-Aware Encoder, the inner workings of the unified reasoning module, and the precise composition of the final loss function.
\section{Experiments}
\label{experiment}
\begin{table*}[h!]
  \caption{Statistical information of datasets.}
  \label{Table: dataset}
  \setlength{\tabcolsep}{4pt}
  \centering
  \begin{tabular}{ccccccc}
    \toprule
    Dataset & \#Entities & \#Relations & \#Time interval & \#Training set & \#Validation set & \#Test set \\
    \midrule
    GDELT & 500 & 20 & 15 mins & 2,735,685 & 341,961 & 341,961 \\
    ICEWS05-15 & 10,488 & 251 & 24 hours & 386,962 & 46,092 & 462,775 \\
    ICEWS18 & 23,033 & 256 & 24 hours & 373,018 & 45,995 & 49,545 \\
    ICEWS14 & 7,128 & 230 & 24 hours & 63,685 & 13,823 & 13,222 \\
    YAGO & 10,623 & 10 & 1 year & 16,406 & 2,050 & 2,051 \\
    WIKI & 12,544 & 24 & 1 year & 32,497 & 4,062 & 4,062 \\
    \bottomrule
  \end{tabular}
\end{table*}
\subsection{Statistics of datasets}
\label{Statistics of datasets}

Table \ref{Table: dataset} presents the statistics of the six temporal knowledge graph datasets used in our experiments. All these datasets follow open-source licenses. We employed datasets with varying temporal granularities, from GDELT's fine-grained 15-minute intervals capturing political events to YAGO and WIKI's yearly intervals representing long-term factual knowledge. The ICEWS family (ICEWS05-15, ICEWS18, ICEWS14) features daily temporal resolution with different entity scales, with ICEWS18 containing the largest number of entities (23,033) and ICEWS05-15 spanning the longest time period. Notable structural differences include GDELT's dense connectivity despite fewer entities (500), contrasting with YAGO's sparse structure (10,623 entities but only 10 relation types). All datasets are divided into training (80\%), validation (10\%), and test sets (10\%). For interpolation reasoning, we split the dataset of GDELT and ICEWS05-15/18 by timestamps. And for extrapolation reasoning, we split the dataset of ICEWS14, YAGO, and WIKI by time intervals.

\subsection{Detailed Experimental Settings}
\label{sec:full_experimental_settings}
Our model is implemented in Python using the PyTorch library and trained on a single NVIDIA RTX 3090 GPU with 24GB of memory. 

For the model architecture, we set the embedding dimension to 320. The dynamic subgraph construction collects a maximum of 50 neighboring facts (max context size). During this process, we apply a structural hop decay factor of $\eta=0.8$, and the temporal fallback mechanism retrieves the 3 nearest neighbors ($k=3$) for sparse regions. The subgraph encoder is a 2-layer SSAE that utilizes ReLU activations and self-loops. The unified reasoning module is built upon a Transformer encoder with 8 layers and 8 attention heads. To regularize the model, the diffusion process uses 1000 steps with a linear noise schedule ranging from $1\times10^{-4}$ to $0.02$, and its denoiser is a 3-layer MLP with a hidden dimension of 256.

For training, we use the Adamax optimizer with a batch size of 256. The learning rate is set to 0.02, with a warmup proportion of 0.1 and a weight decay of 0.01. A context dropout rate of 0.3 is applied to the neighborhood information during training. The relative weight for the diffusion prediction loss of $\lambda_{\text{diff}}==1.0$ and a relative weight for the time prediction loss of $\lambda_{\text{time}}=2.0$.

\subsection{Baselines}
\label{baselines}
\textbf{Extrapolation reasoning} in temporal knowledge graphs involves predicting future facts based on historical data. The methods can be categorized as follows:

\textbf{GNN-based Methods for Extrapolation Reasoning}
\begin{itemize}
    \item \textbf{RE-GCN} is a recurrent evolution network based on graph convolution that learns evolutionary representations of entities and relations.
    \item \textbf{L2TKG} is a latent relations learning approach that mines missing associations between entities, focusing on both intra-time relations and inter-time relations to improve prediction accuracy.
    \item \textbf{DREAM} is an adaptive reinforcement learning model that uses attention mechanisms to simultaneously capture temporal evolution and semantic dependencies.
    \item \textbf{LogCL} uses entity-aware attention mechanisms to capture query-relevant information in both local and global historical contexts, while employing contrastive learning to guide their fusion and enhance noise resistance. 
    \item \textbf{DyMemR} implements a human-inspired memory pool to retain relevant historical information.
    \item \textbf{THCN} incorporates causality with dilation factors to capture temporal dependencies.
\end{itemize}

\textbf{Other Methods for Extrapolation Reasoning}
\begin{itemize}
    \item \textbf{CSI} extracts causal sub-histories while eliminating spurious correlations to better represent the logical foundations of predictions.
    \item \textbf{TempValid} models rule confidence as time-dependent rather than static by learning both confidence and decay coefficients.
    \item \textbf{LLMDA} dynamically updates LLM-generated rules with emerging events, enabling adaptation to evolving dynamics within temporal knowledge graphs.
\end{itemize}

\textbf{Interpolation reasoning} in knowledge graphs involves predicting missing facts within the known temporal range. The methods can be categorized as follows:

\begin{itemize}
    \item \textbf{RotateQVS} employs spatial rotation operations around the time axis to represent temporal relationships, offering an innovative approach to modeling time-dependent interactions.
    \item \textbf{TeAST} maps relations onto Archimedean spiral timelines in complex space rather than incorporating temporal information into entities.
    \item \textbf{HYIE} further enriches temporal representation by learning structures across Euclidean, hyperbolic, and hyper-spherical spaces.
    \item \textbf{LCGE} jointly models the time-sensitive aspects of events, including both timeliness and causality.
    \item \textbf{HGE} embeds temporal knowledge graphs in a product space combining complex, split-complex, and dual geometric subspaces with temporal-relational and temporal-geometric attention mechanisms to effectively model diverse temporal patterns that cannot be captured by single-space embeddings.
    \item \textbf{5EL} integrates Large Language Models with Möbius Group transformations on the Riemann sphere to effectively model complex temporal patterns in knowledge graphs.
    \item \textbf{TGeomE} represents entities, relations, and time as multivectors, providing superior expressiveness by subsuming several state-of-the-art models while achieving better performance through geometric product operations and a novel linear temporal regularization..
\end{itemize}

\textbf{Unified Methods.} These methods are applicable to both extrapolation and interpolation reasoning:

\begin{itemize}
    \item \textbf{TPAR} uses a recursive encoding method based on the Bellman-Ford algorithm to score destination entities along temporal paths, providing better performance, robustness, and interpretability compared to methods designed for only one reasoning setting.
    \item \textbf{ECEformer} is a Transformer-based model that learns evolutionary chains of events by combining a Transformer encoder to capture intra-quadruple semantics with MLP-based mixing layers for inter-quadruple relationships.
\end{itemize}

\subsection{Case Study}
\label{Case study}

To qualitatively assess DynaGen's reasoning capabilities, we conduct case studies on representative examples for both interpolation and extrapolation tasks. By examining the model's top-ranked predictions, we can gain deeper insights into its behavior beyond aggregate metrics.

First, we examine interpolation reasoning on the \textbf{YAGO} dataset. The results, shown in Table \ref{tab:interpolation_case}, demonstrate the model's prediction capabilities on two distinct queries.

\begin{table*}[h!]
  \caption{Interpolation Reasoning Case Study on YAGO: Ranking of Subject and Object Entity Predictions.}
  \label{tab:interpolation_case}
  \centering
  \small

  \begin{tabular}{ccc|ccc}
    \toprule
    \multicolumn{3}{c|}{Case 1: [S] 88 [P] 4 [O] 1668 (2006, 2006)} & \multicolumn{3}{c}{Case 2: [S] 349 [P] 8 [O] 7694 (2000, 2000)} \\
    \midrule
    \multicolumn{1}{c}{Rank} & \multicolumn{1}{c}{Subject (score)} & \multicolumn{1}{c|}{Object (score)} & 
    \multicolumn{1}{c}{Rank} & \multicolumn{1}{c}{Subject (score)} & \multicolumn{1}{c}{Object (score)} \\
    \midrule
    1 & \textbf{88 (20.062)} & \textbf{1668 (21.701)} & 1 & 7693 (11.250) & 7778 (10.462) \\
    2 & 1668 (18.480) & 88 (14.579) & 2 & 7793 (10.354) & 7708 (10.294) \\
    3 & 89 (15.783) & 1181 (13.119) & 3 & 278 (10.248) & 7730 (9.651) \\
    4 & 2742 (13.887) & 766 (12.860) & 4 & 7707 (9.229) & \textbf{7694 (9.243)} \\
    5 & 1181 (13.380) & 3380 (12.542) & 5 & \textbf{349 (9.044)} & 7768 (8.231) \\
    6 & 486 (13.278) & 2429 (11.940) & 6 & 271 (8.294) & 3894 (5.979) \\
    7 & 3367 (13.231) & 1096 (11.465) & 7 & 3684 (7.748) & 4838 (5.652) \\
    8 & 1981 (13.055) & 486 (11.410) & 8 & 1780 (7.245) & 7779 (5.613) \\
    9 & 3507 (12.871) & 89 (11.256) & 9 & 389 (6.656) & 5602 (5.434) \\
    10 & 3547 (12.763) & 2390 (11.144) & 10 & 6616 (5.754) & 5503 (5.184) \\
    \bottomrule
  \end{tabular}
  
\end{table*}

In \textbf{Case 1}, for the query `([S] 88, [P] 4, [O] 1668)` in 2006, our model achieved optimal performance by ranking both the correct subject (88) and object (1668) in the \textbf{first position} with high confidence scores. This highlights the model's ability to accurately capture and reason over established temporal patterns.

In \textbf{Case 2}, for the query `([S] 349, [P] 8, [O] 7694)` in 2000, performance remained strong, with the correct subject (349) ranked \textbf{fifth} and the object (7694) ranked \textbf{fourth}. Despite not being the top result, both ground-truth entities appeared within the top five predictions, confirming the model's robust reasoning capabilities across different temporal contexts and demonstrating its effectiveness even on less frequent or more complex relationships.

Next, we evaluate extrapolation reasoning on the \textbf{ICEWS05-15} dataset, with results presented in Table \ref{tab:extrapolation_case}. This tests the model's ability to generalize to previously unseen time periods.

\begin{table*}[h!]
  \caption{Extrapolation Case Study on ICEWS05-15: Ranking of Subject and Object Predictions.}
  \label{tab:extrapolation_case}
  \centering
  \small
  \begin{tabular}{ccc}
    \toprule
    \multicolumn{3}{c}{Case 3: [S] Police (Sudan) [P] Arrest [O] Citizen (Sudan) [T] 2015-05-21} \\
    \cmidrule(lr){2-2} \cmidrule(l){3-3}
    \multicolumn{1}{c}{Rank} & \multicolumn{1}{c}{Subject (score)} & \multicolumn{1}{c}{Object (score)} \\
    \midrule
    1 & \textbf{Police (Sudan) (7.451)} & \textbf{Citizen (Sudan) (7.837)} \\
    2 & Sudan (5.683) & Criminal (Sudan) (6.522) \\
    3 & Military Personnel - Special (Sudan) (5.097) & Children (Sudan) (5.450) \\
    4 & Special Court (Sudan) (4.504) & Armed Opposition (Sudan) (3.911) \\
    5 & Government (Sudan) (3.786) & Sudan (3.626) \\
    \midrule
    \multicolumn{3}{c}{Case 4: [S] China [P] Express intent to meet [O] Japan [T] 2007-03-20} \\
    \cmidrule(lr){2-2} \cmidrule(l){3-3}
    \multicolumn{1}{c}{Rank} & \multicolumn{1}{c}{Subject (score)} & \multicolumn{1}{c}{Object (score)} \\
    \midrule
    1 & \textbf{China (3.804)} & \textbf{Japan (4.675)} \\
    2 & South Korea (3.693) & South Korea (4.234) \\
    3 & Wen Jiabao (1.718) & France (2.399) \\
    4 & Roh Moo Hyun (1.632) & Thailand (2.161) \\
    5 & Alberto Fujimori (0.757) & Kazakhstan (2.117) \\
    \bottomrule
  \end{tabular}
\end{table*}

In \textbf{Case 3}, for the event `(Police (Sudan), Arrest, Citizen (Sudan))` on 2015-05-21, the model exhibited exceptional extrapolation performance. It ranked both the correct subject and object as \textbf{first}, with robust confidence scores. This demonstrates the model's ability to effectively project recurring sociopolitical patterns into future, unseen time points.

Similarly, for \textbf{Case 4}, concerning the diplomatic event `(China, Express intent to meet, Japan)` on 2007-03-20, the model again achieved excellent accuracy, ranking both \textbf{China} and \textbf{Japan} first. The variation in score magnitudes between the cases (higher for the common law enforcement event, lower for the more variable diplomatic event) reveals how the model's confidence reflects the frequency and predictability of event patterns. Notably, in both cases, the model identified geopolitically plausible alternatives in the top rankings (e.g., \textit{South Korea}, \textit{Wen Jiabao}), illustrating its ability to capture meaningful semantic relationships when reasoning about the future.

In summary, these case studies offer qualitative evidence that complements our quantitative benchmarks. By examining interpolation on YAGO and extrapolation on ICEWS05-15, we not only confirm DynaGen’s capacity to rank correct entities highly but also illuminate its robustness in handling both frequent and rare event patterns. These examples highlight the model's sensitivity to temporal context and underscore its generalization to unseen time points, substantiating its practical utility for real-world temporal reasoning tasks.

\subsection{Training Overhead and Inference Efficiency}
Although DynaGen's complex architecture leads to significant training overhead, its inference process is highly efficient, requiring only 4 seconds for a single test batch (batch size is 256) on the YAGO dataset. This level of performance is more than sufficient for applications in temporal knowledge graph reasoning, which typically do not have strict real-time requirements.

\section{Methodological Details}

\subsection{Temporal Subgraph Construction}
\label{sec:appendix_subgraph_algo}
The construction of entity-centric temporal subgraphs is guided by a Breadth-First Search (BFS) algorithm initiated from a query entity $e_i$ at time $t_i$. The algorithm expands outwards for a predefined number of hops, collecting neighboring facts that fall within a dynamically determined temporal window. This process is augmented with several mechanisms to ensure robustness and contextual relevance.

\subsubsection{Fallback for Temporal Sparsity}
For entities located in temporally sparse regions of the graph where no neighbors are found within the initial time window, a fallback strategy is activated. Instead of generating an empty subgraph, the model identifies a small number of neighbors ($k$) with the minimum temporal distance to the query time $t_i$. This ensures that every entity is associated with a meaningful local context for subsequent encoding.

\subsubsection{Structural and Temporal Edge Weighting}
The significance of each edge in the subgraph is quantified by a composite weight that reflects both temporal proximity and structural distance. Each edge $e$ is assigned a temporal weight based on an exponential decay function of the time difference between the edge's timestamp $t_e$ and the query time $t_i$:
\begin{equation}
w_{\text{time}}(e) = \exp(-\gamma|t_i - t_e|)
\end{equation}
where $\gamma$ is a learnable decay parameter. This temporal weight is then modulated by a structural decay factor $\eta^h$, where $h$ is the hop distance from the query entity and $\eta$ is a fixed hyperparameter. The final edge weight, $w'(e) = \eta^h \cdot w_{\text{time}}(e)$, prioritizes facts that are both recent and structurally close to the query entity.

\subsubsection{Handling of Isolated Entities}
In the rare case that an entity is completely isolated, a minimal subgraph containing only the entity itself and a self-loop is constructed. This ensures the batch-processing pipeline remains intact and allows the model to generate a base representation from the entity's own features.

\subsubsection{Adaptability to Static Graphs}
The framework is designed to be backward-compatible with static (non-temporal) knowledge graphs. When a query lacks a timestamp, the temporal filtering logic is bypassed, and the model constructs a standard 1-hop ego-graph, including all immediate neighbors and their inverse relations.

\subsection{Synergistic Structure-Aware Encoder}
\label{sec:appendix_SSAE}
The SSAE module refines node representations within the constructed subgraphs by synergistically integrating semantic and structural information.

\subsubsection{Layer Connectivity and Residual Updates}
The encoder consists of multiple stacked blocks. Within each block, a node's representation is processed in parallel by two branches: a Relational Graph Convolutional Network to capture relation-specific semantics and a Graph Attention Network to weigh structural importance. The outputs from both branches are concatenated and fused via a linear transformation. To facilitate the training of a deeper architecture and mitigate vanishing gradients, we employ residual connections. The fused message is integrated with the node's representation from the previous layer, followed by a ReLU activation, to produce the updated representation for the next block.

\subsubsection{Node-level Temporal Gating}
To efficiently incorporate temporal information into the GNN updates, we employ a node-level gating mechanism. For each node $v$ in the subgraph, its aggregated message $m_v^{(fused)}$ is concatenated with a scalar feature representing the average temporal-structural weight of its incident edges. This combined vector is then passed through a small feed-forward network with a sigmoid activation to produce a gating vector $g_v \in \mathbb{R}^d$. The final message is modulated via an element-wise product with this gate, $m_v' = g_v \odot m_v^{(fused)}$. This approach effectively infuses temporal awareness by scaling node features based on the overall recency of their local neighborhood, while being more computationally efficient than an edge-level alternative.

\subsection{Unified Reasoning Module}
\label{sec:appendix_reasoning_module}
The unified reasoning module processes the GNN output to produce the final query representation. This design is summarized in Algorithm~\ref{alg:unified_reasoning}.
\begin{algorithm}[h!]
\caption{Unified Reasoning over the Local Neighborhood}
\label{alg:unified_reasoning}
\begin{algorithmic}[1]
\Require Query $(e_i, p_i, t_i)$, GNN output $\mathbf{z}_i$, 1-hop context $\mathcal{C}\!=\!\{(e_i,p_i,t_i)\}_{i=1}^k$, Positional embeddings $\mathbf{P}_j$.
\Ensure Final representation $\mathbf{z}_{\text{final}}$.

\State // 1. Construct input sequence with masked timestamp for the query
\State $\mathbf{C}_0 \gets [\texttt{[CLS]}, \mathbf{z}_{i}, \mathbf{p}_i, \texttt{[MASK]}]$
\State $\mathcal{S} \gets (\mathbf{C}_0, \mathcal{C}_1, \dots, \mathcal{C}_k)$

\State // 2. Add positional embeddings to each element in the sequence
\For{$i=0$ \textbf{to} $k$}
    \State $\mathbf{S}'_i \gets [\mathbf{S}_{i,j} + \mathbf{P}_j]_{j=0}^3$  \Comment{e.g., for [ent, rel, time, ...]}
\EndFor

\State // 3. Stage 1: Atomic Transformer encoding
\State $\mathbf{H} \gets \texttt{TransformerEncoder}(\{\mathbf{S}'_i\}_{i=0}^k)$

\State // 4. Stage 2: Global MLP-Mixer aggregation
\State $\mathbf{h}_{\text{mix}} \gets \text{Flatten}(\mathbf{H})$
\For{$l=1$ \textbf{to} $L/2$}
    \State $\mathbf{h}_{\text{mix}} \gets \mathbf{h}_{\text{mix}} + \text{MLP}_{\text{token}}(\text{LayerNorm}(\mathbf{h}_{\text{mix}}))$
    \State $\mathbf{h}_{\text{mix}} \gets \mathbf{h}_{\text{mix}} + \text{MLP}_{\text{channel}}(\text{LayerNorm}(\mathbf{h}_{\text{mix}}))$
\EndFor

\State // 5. Final pooling and output
\State $\mathbf{z}_{\text{final}} \gets \text{MeanPool}(\mathbf{h}_{\text{mix}})$
\State \Return $\mathbf{z}_{\text{final}}$
\end{algorithmic}
\end{algorithm}

\subsubsection{Contextual Timestamp Masking}
To enhance the model's understanding of temporal context, we employ a self-supervised auxiliary task similar to Masked Language Modeling (MLM). During training, we randomly select facts from the 1-hop neighborhood context of a query. With a certain probability, the time embedding of a selected fact is corrupted, either by replacing it with a special learnable \texttt{[MASK]} token or with the embedding of a randomly sampled timestamp. The model is then trained to predict the original timestamp from the final aggregated representation of the context. This task, which contributes the $\mathcal{L}_{\text{time}}$ loss component, compels the model to learn robust temporal dependencies within the local graph structure.

\subsubsection{Atomic Positional Encoding}
To enable the Transformer-based encoder to distinguish between the components of a temporal fact (subject, predicate, object, time), we use a dedicated, learnable atomic positional encoding. Four unique positional vectors are learned, corresponding to the four slots in an augmented fact tuple (e.g., [CLS], entity, relation, time). These vectors are added to the feature embeddings of the corresponding components before they are fed into the self-attention layers, thereby explicitly informing the model of each element's structural role.

\subsection{Composition of the Final Loss Function}
\label{sec:appendix_loss}
The model is trained end-to-end by optimizing a composite loss function. The total loss is a weighted sum of a primary prediction loss and several auxiliary objectives designed to regularize and guide the learning of different model components. The final loss function is defined as:
\begin{equation}
\mathcal{L} = \mathcal{L}_{\text{pred}} + \lambda_{\text{diff}}\mathcal{L}_{\text{diff}} + \lambda_{\text{time}} \mathcal{L}_{\text{time}}
\end{equation}
where the components are:
\begin{itemize}
    \item $\mathcal{L}_{\text{pred}}$: The primary entity prediction loss, calculated from the final output of the unified reasoning module.
    \item $\mathcal{L}_{\text{time}}$: The self-supervised loss from the contextual timestamp masking task.
    \item $\mathcal{L}_{\text{diff}}$: The mean-squared error loss from the diffusion regularizer, which encourages the GNN to learn robust representations.
\end{itemize}
The hyperparameters $\lambda_{\text{diff}}$ and $\lambda_{\text{time}}$ balance the influence of the auxiliary tasks. This multi-task framework ensures that all modules are jointly optimized.

\end{document}